\documentclass[letterpaper]{article} 

\usepackage{aaai2026}  
\usepackage{times}  
\usepackage{helvet}  
\usepackage{courier}  
\usepackage[hyphens]{url}  
\usepackage{graphicx} 
\usepackage{natbib}  
\usepackage{caption} 
\usepackage{adjustbox}
\usepackage{bbding}
\usepackage{amssymb}
\usepackage{booktabs,tabularx}
\usepackage{amsmath}
\usepackage{subcaption}

\usepackage[pagebackref,breaklinks]{hyperref}
\definecolor{citecolor}{HTML}{3498DB}
\hypersetup{colorlinks,linkcolor={red},citecolor={citecolor},urlcolor={black}} 
\usepackage[capitalize]{cleveref}

\usepackage{algorithm}
\usepackage{algorithmic}

\usepackage{newfloat}

\usepackage{listings}
\DeclareCaptionStyle{ruled}{labelfont=normalfont,labelsep=colon,strut=off} 
\floatstyle{ruled}
\newfloat{listing}{tb}{lst}{}
\floatname{listing}{Listing}
\title{OmniTraj: Pre-Training on Heterogeneous Data for Adaptive and Zero-Shot Human Trajectory Prediction}
\author{Yang Gao, Po-Chien Luan, Kaouther Messaoud, Lan Feng, Alexandre Alahi}

\affiliations{
    Visual Intelligence for Transportation (VITA) laboratory\\
    EPFL, Switzerland\\
    \texttt{\{firstname.lastname\}@epfl.ch} \\

}

\begin{document}
\nocopyright 

\maketitle

\begin{abstract}

While large-scale pre-training has advanced human trajectory prediction, a critical challenge remains: zero-shot transfer to unseen dataset with varying temporal dynamics. State-of-the-art pre-trained models often require fine-tuning to adapt to new datasets with different frame rates or observation horizons, limiting their scalability and practical utility. In this work, we systematically investigate this limitation and propose a robust solution. We first demonstrate that existing data-aware discrete models struggle when transferred to new scenarios with shifted temporal setups. We then isolate the temporal generalization from dataset shift, revealing that a simple, explicit conditioning mechanism for temporal metadata is a highly effective solution. Based on this insight, we present OmniTraj, a Transformer-based model pre-trained on a large-scale, heterogeneous dataset. Our experiments show that explicitly conditioning on the frame rate enables OmniTraj to achieve state-of-the-art zero-shot transfer performance, reducing prediction error by over 70\% in challenging cross-setup scenarios. After fine-tuning, OmniTraj achieves state-of-the-art results on four datasets, including NBA, JTA, WorldPose, and ETH-UCY. The code is publicly available: \href{https://github.com/vita-epfl/omnitraj}{\color{magenta}https://github.com/vita-epfl/omnitraj}.

\end{abstract}    
\section{Introduction}
\label{sec:intro}

Human trajectory prediction is a critical component in applications such as autonomous driving, robotics, and surveillance. However, existing models struggle to generalize across datasets with varying temporal dynamics, requiring extensive fine-tuning to adapt to new environments. This limitation hinders practical deployment, especially in scenarios where labeled data is scarce.
While fine-tuning on target-domain data improves accuracy by adapting the model to specific scenarios, it remains labor-intensive and dataset-dependent. In contrast, zero-shot transfer offers a scalable solution by leveraging pre-trained representations to predict trajectories without additional training. In this paper, we tackle a key challenge: zero-shot transfer to unseen datasets with varying temporal dynamics, such as different frame rates and observation lengths.

\begin{figure}[htbp]
    \centering
    \includegraphics[width=\linewidth]{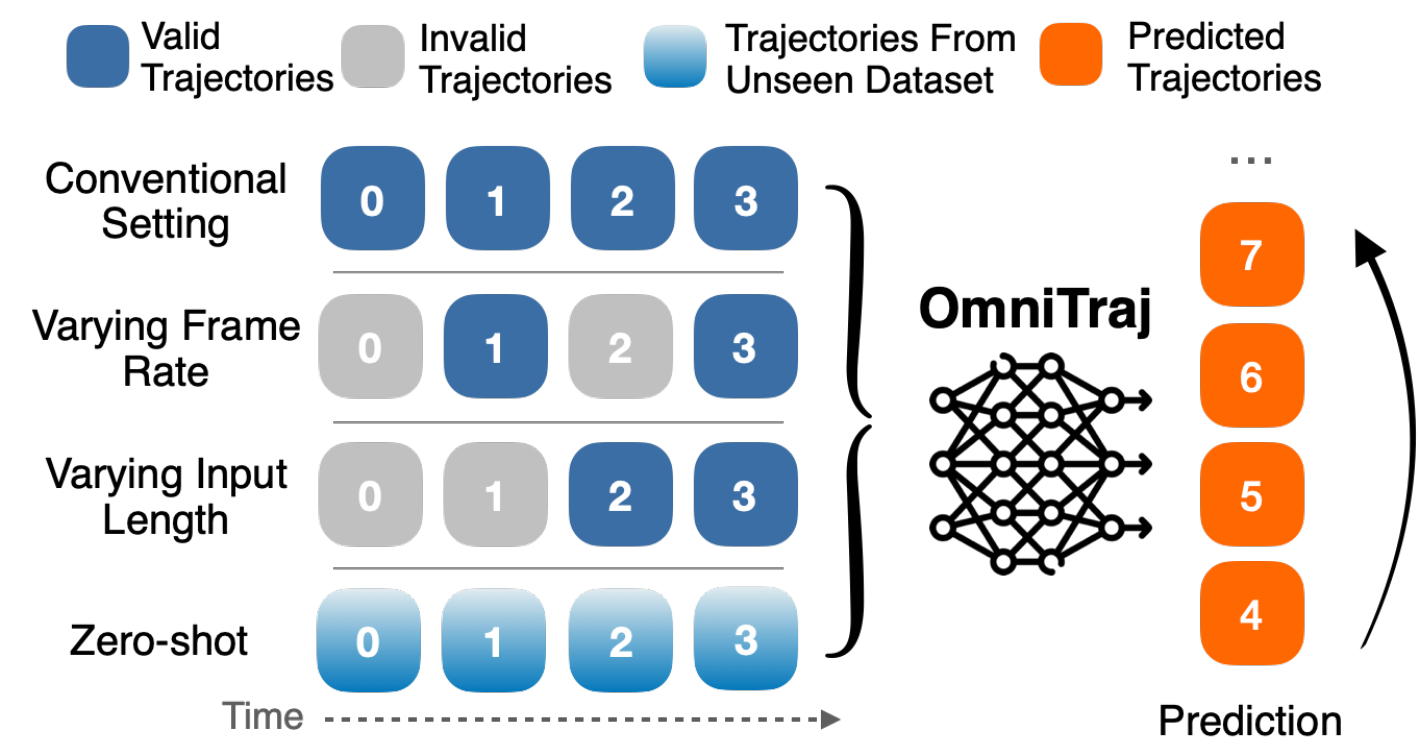}
    \caption{\textbf{OmniTraj: A pre-trained trajectory predictor that adapts to varying frame rates and horizons while excelling in zero-shot transfer.} }
    \label{fig:model_overview}
\end{figure}

Conventional trajectory prediction models~\cite{alahi2016social,girgis2022latent} remain constrained by assumptions that limit their generalizability across scenarios since most approaches handle the input and output with fixed observation and prediction horizons at a specific frame rate, making them inflexible when adapting to new scenarios with different frame setups, as shown in~\Cref{fig:model_overview}.
Recent advances in large-scale pre-training have sought to address this generalization challenge. Models like Multi-Transmotion~\cite{gao2024multi} leverage massive, diverse datasets to learn robust representations via spatial-temporal masking. However, despite being pre-trained on varied data, they still rely on a fixed temporal resolution and require fine-tuning to adapt to new scenarios with different frame setups. This highlights a critical research gap: current pre-training paradigms do not inherently solve the problem of zero-shot temporal generalization. While alternative approaches, such as continuous-time models like TrajSDE~\cite{park2024improving}, aim to be frame-rate agnostic, the most effective strategy for zero-shot transfer remains an open question.

In this work, we systematically investigate the challenge of zero-shot temporal generalization. We first quantify the performance degradation of pre-trained models when transferred to unseen datasets with different frame setups. We then design an experiment to isolate the effect of temporal generalization from dataset shift, allowing us to directly compare different strategies for handling temporal variance. Our findings reveal that a simple and efficient mechanism—explicitly conditioning the model on temporal metadata—is a remarkably effective solution. Based on this insight, we introduce OmniTraj, as shown in ~\Cref{fig:model_overview}, a flexible Transformer-based model built upon a large-scale, heterogeneous data framework that leverages an FPS-aware design to achieve robust zero-shot and few-shot performance.

To achieve this, we consider three key aspects: data, model, and training. From a data perspective, a robust trajectory prediction model requires a large, flexible dataset. We design a frame rate- and horizon-agnostic data container that aligns diverse trajectories with meta-information like FPS and prediction horizon. We extend this framework to \textbf{859 hours} of motion data from \textbf{12 diverse datasets}, creating a large-scale, multi-modal dataset invariant to frame rate and horizon length. To fully leverage this diverse data, we introduce a Transformer-based architecture specifically to handle such heterogeneity. The key innovations are our use of the \textbf{FPS-aware embeddings} and the \textbf{Decoupled interaction modules}, which explicitly informs the model of the temporal resolution of the input data and strengthens the social interaction between agents. For training, we employ the masked multi-modal pre-training and compare it with trajectory-only pre-training to investigate how it affects the cross-domain transfer of the model. Since trajectory-only data is more widely available than visual modalities like 3D pose, we prioritize it at inference, ensuring greater applicability in real-world settings.

This work makes three main contributions. 
\begin{itemize}
    \item We provide the first systematic analysis of zero-shot transfer on unseen datasets with varying temporal dynamics for pre-trained trajectory predictors. We demonstrate via isolated experiments that an explicit conditioning mechanism is more robust for this task than data-unaware discrete models or continuous-time approaches.
    \item We introduce OmniTraj, a pre-trained model incorporating our findings. It reduces zero-shot transfer by over 70\% and achieves state-of-the-art performances on four datasets: NBA, JTA, WorldPose, and ETH-UCY.
    \item To facilitate this research, we develop and will release UniHuMotion++, the largest unified data framework with native support for heterogeneous temporal setups in the human motion prediction area.
    
\end{itemize}

\section{Related Work}
\label{sec:formatting}

\textbf{Conventional Human trajectory prediction} involves forecasting future motion given a fixed-length observation of past trajectories~\cite{rudenko2020human}. A wide range of approaches have been explored, with traditional methods focusing on modeling social interactions through techniques such as social pooling~\cite{alahi2016social,kothari2021human}, Graph Neural Networks~\cite{huang2019stgat, li2020evolvegraph,salzmann2020trajectron++, xu2022groupnet, mohamed2020social}, and attention mechanisms~\cite{messaoud2020attention, giuliari2021transformer, yuan2021agentformer, Girgis2021LatentPrediction}. In parallel, generative models—including Generative Adversarial Networks~\cite{gupta2018social, sadeghian2019sophie, kosaraju2019social}, Variational Autoencoders~\cite{schmerling2018multimodal, ivanovic2020multimodal, xu2022socialvae}, and diffusion models~\cite{gu2022trajdiffusion, mao2023leapfrog, bae2024singulartrajectory}—have been investigated to capture multi-modal future distributions. More recently, human pose has been identified as a strong visual cue, enhancing trajectory prediction performance~\cite{saadatnejad2024socialtransmotion} or augmenting unified localization and trajectory prediction~\cite{luan2025unified}. However, these methods are constrained by fixed data scenarios and temporal settings, limiting their ability to generalize across diverse scenarios.

\noindent\textbf{Pre-training for trajectory prediction} involves leveraging self-supervised learning to pre-train models to achieve better generalization and robustness. Relevant methods are designing self-supervised tasks such as contrastive learning~\cite{liu2021social, bhattacharyya2023ssl, xu2022pretram} and 
masking~\cite{chen2023traj, lan2023sept, cheng2023forecast, gao2024multi}. However, these works still require fine-tuning on samples to adapt to different datasets, struggling to attain zero-shot knowledge transfer.

\noindent\textbf{Unified data framework} is crucial for developing generalized trajectory prediction models. Several efforts have sought to standardize datasets and evaluation protocols. TrajData~\cite{ivanovic2023trajdata} introduces a unified interface for trajectory and map data, simplifying model training and evaluation. UniTraj~\cite{Feng2024} extends this idea by standardizing datasets, models, and evaluation metrics, demonstrating that increasing data diversity significantly enhances performance. UniHuMotion~\cite{gao2024multi} integrates seven diverse datasets and proposes a Transformer-based model with a masking strategy for cross-modal pretraining, achieving strong results across multiple trajectory prediction tasks. Despite these advancements, existing approaches rely on fixed temporal horizons, which is impractical for real-world applications where observation durations vary.

\noindent\textbf{Dynamic frame setups} are essential to be handled by a pre-trained model in human trajectory prediction. In real-world scenarios, observation lengths may be insufficient, and frame rates can vary across different systems due to hardware constraints. Prior works have explored trajectory prediction under limited observations. MOE~\cite{sun2022human} and DTO~\cite{monti2022many} investigate models that rely on only two historical frames for accurate forecasting. To address missing frames, BCDiff~\cite{li2023bcdiff} employs a bi-directional diffusion framework to iteratively refine both future and unobserved past trajectories. FLN \cite{xu2024adapting} mitigates performance degradation caused by varying observation lengths using temporally invariant representations and adaptive refinement. TrajSDE~\cite{park2024improving} uses a continuous-time model to make the model frame-rate agnostic and increase the generalization. Recently, LaKD~\cite{li2025lakd} enables trajectory prediction from arbitrary observation lengths by dynamically exchanging information among trajectories. However, these methods do not integrate varying frame rates in the efficient discrete models, which hinders the development of pre-trained models with zero-shot transfer. Our work aims to address these limitations by not only unifying the data framework but also developing a model capable of handling diverse frame rates and horizons, moving toward a more generic pre-trained model with strong zero-shot transfer ability.

\section{Unified Human Motion Data Framework (UniHuMotion++)}
\label{sec:data_framework}

One of the key challenges in pre-training a generic model with zero-shot ability is the availability of a comprehensive and diverse dataset. UniHuMotion~\cite{gao2024multi} introduced a framework that unifies multiple human motion datasets by standardizing data formats and fixing frame setups. However, this framework remains insufficient for developing a pre-trained model that can deal with unseen scenarios featuring different frame setups. To address this, a more flexible data framework is required—one that can efficiently handle varying frame rates and prediction horizons, allowing the model to incorporate a broader range of motion data during training.

Building upon the initial UniHuMotion framework~\cite{gao2024multi}, we introduce UniHuMotion++, an enhanced version that can combine datasets with different frame setups into one framework. We also integrate five additional datasets and improve data writing and storage strategies. These upgrades enable the framework to support a larger variety of datasets with different frame setups while caching data more efficiently. The UniHuMotion++ currently contains 859 hours human motion of 12 datasets, including: NBA Sport VU~\cite{Nba2016}, JRDB-Pose~\cite{vendrow2023jrdbpose}, JTA~\cite{fabbri2018learning}, Human3.6M~\cite{ionescu2013human3}, AMASS~\cite{mahmood2019amass}, 3DPW~\cite{von2018recovering}, Nuscenes~\cite{caesar2020nuscenes}, WOMD~\cite{ettinger2021large}, Argoverse2~\cite{wilson2023argoverse}, WorldPose~\cite{jiang2024worldpose}, SDD~\cite{robicquet2016learning}, Trajnet++~\cite{kothari2021human}. The detailed information about data horizon, FPS, size (hours), and modalities (e.g., trajectory, 3D/2D bounding box, 3D/2D pose keypoints) can be found in \Cref{tab:UniHuMotion} of our supplementary material.

\section{Method}
\label{sec:method}

In this section, we describe the problem formulation and network architecture of our OmniTraj, which is designed for robust human trajectory prediction across diverse settings.

\paragraph{Problem Formulation.}
We denote the trajectory sequence of agent \( i \) as \( x^T_i \), the 3D and 2D local pose sequences as \( x^{3dP}_i \) and \( x^{2dP}_i \), and the 3D and 2D bounding box sequences as \( x^{3dB}_i \) and \( x^{2dB}_i \), respectively. The observed time-steps are defined as \( t = 1, \ldots, T_{\text{obs}} \) and the prediction time-steps as \( t = T_{\text{obs}}+1, \ldots, T_{\text{pred}} \). In a scene with \( N \) agents, the network input is 
\[
X = \left[ X_1, X_2, \dots, X_N \right],
\]
where for each agent
\[
X_i = \{ x^c_i \mid c \in \{T,\, 3dP,\, 2dP,\, 3dB,\, 2dB\} \}.
\]
Each tensor \( x^c_i \in \mathbb{R}^{T_{\text{obs}} \times e_c \times f_c} \) has \( e_c \) elements (e.g., keypoints) and \( f_c \) features per element. Without loss of generality, we consider \( X_1 \) as the primary agent, and the network's output \( Y = Y_1 \) contains its predicted future trajectory.

\paragraph{Input Cues Embeddings.}
To exploit the diverse visual cues, OmniTraj uses cue-specific embedding layers. For each cue \( c \) and agent \( i \), the raw input \( x^c_i \) is embedded via a Multi-Layer Perceptron (MLP):
\[
H^c_i = \text{MLP}_c(x^c_i) + P,
\]
where \( P \) is the positional encoding capturing the temporal order of the observations. In addition, separate embeddings for person identity and keypoint type are learned to differentiate between agents and to encode keypoint-specific movement information.

\paragraph{Frame Rate Encoding.}  
Since the observation sequences come with different frame rates, we incorporate a dedicated frame rate encoding module. We handle a frame rate \( r \) by:
\[
E_r = \text{MLP}_r(r),
\]
which converts the scalar \( r \) into a latent embedding. This frame rate embedding is then integrated with the multi-modal tokens element-wise addition, 
normalizing the temporal resolution across inputs. This process allows OmniTraj to consistently interpret motion dynamics regardless of whether the data is captured at high or low frame rates. As detailed in \Cref{tab:fps_encoder_design}, we explored alternative encoder designs and compared them against state-of-the-art models to validate their effectiveness for cross-domain generalization.

\begin{figure*}[!htbp]
    \centering
    \includegraphics[width=\linewidth]{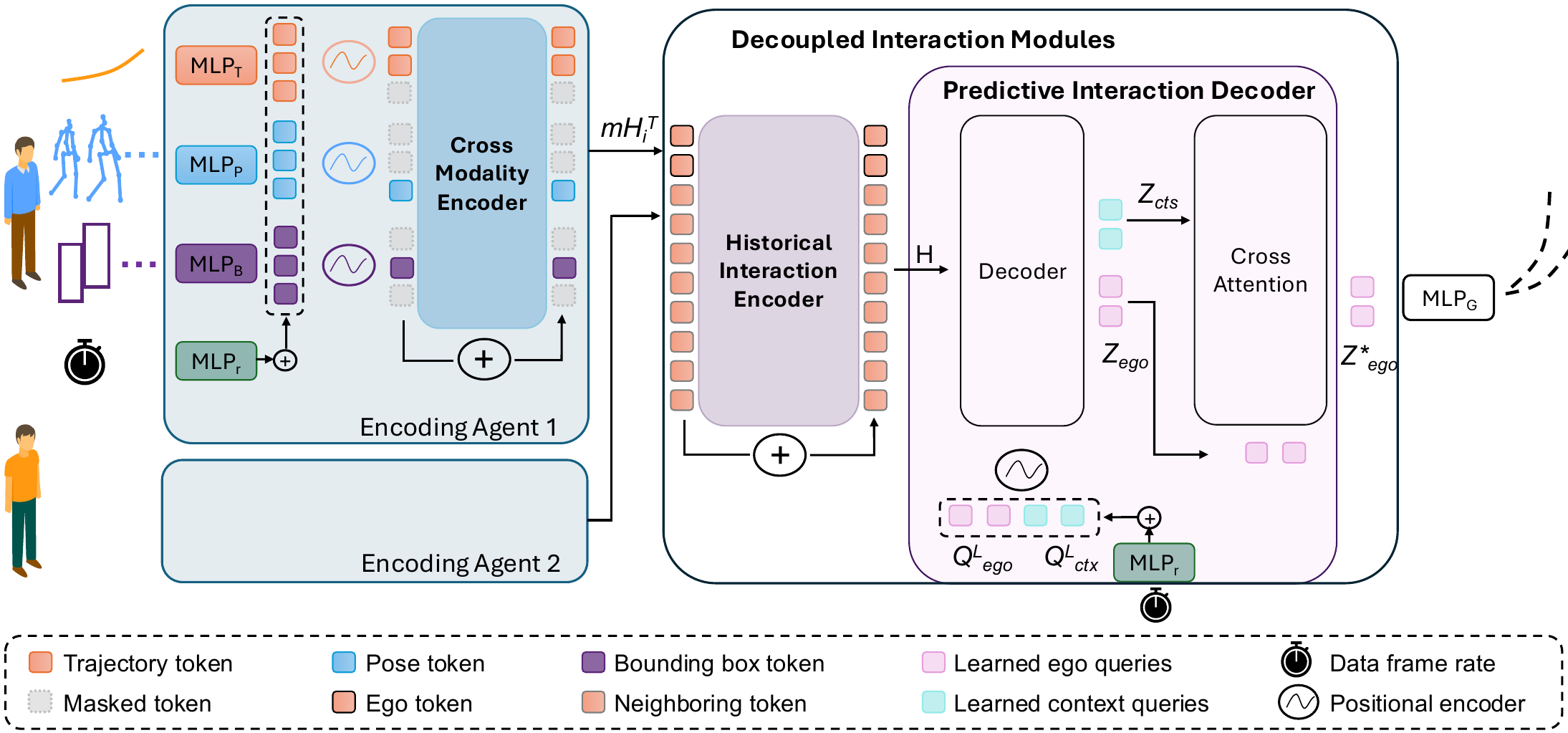}
    \caption{\textbf{OmniTraj: a decoupled interactive transformer with explicit frame-rate awareness.} The model applies an FPS embedding to input tokens for temporal conditioning. The decoupled design first uses the Historical Interaction Encoder (HIE) to process observed social dynamics. Then, the Predictive Interaction Decoder (PID) models future interactions using learned queries and an ego-centric cross-attention mechanism to refine the diverse future trajectory.}
    \label{fig:model}
\end{figure*}

\paragraph{Cross-Modality Encoder (CME)}
In OmniTraj, the CME is responsible for processing and fusing input embeddings from various modalities. Rather than relying on separate input queries, CME leverages trajectory-associated tokens—directly derived from the agent's observed trajectory—to form a rich, motion-centric representation. This design enables the model to focus on the intrinsic dynamics of the agent while incorporating complementary cues.

Let \( H^T_i \in \mathbb{R}^{T_{\text{obs}} \times D_T} \) denote the embedding tensor for the trajectory modality of agent \( i \), where \( T_{\text{obs}} \) is the number of observed time-steps and \( D_T \) is the feature dimension for the trajectory. In addition, for each auxiliary cue \( c \in \{3dP, 2dP, 3dB, 2dB\} \), the corresponding embedding tensor \( H^c_i \in \mathbb{R}^{T_{\text{obs}} \times e_c \times D_c} \) is computed, with \( e_c \) and \( D_c \) representing the number of elements and the feature dimension for cue \( c \), respectively.

CME employs shared parameters to process these modality-specific embeddings, ensuring a consistent encoding scheme across different inputs. The trajectory embeddings \( H^T_i \) are transformed into refined motion tokens, denoted by \( mH^T_i \), which capture the core dynamics of the agent’s motion. Simultaneously, the auxiliary cues \( H^c_i \) are individually processed and mapped to a common latent space, yielding transformed tensors \( mH^c_i \).

\paragraph{Decoupled Interaction Modules}
To capture complex motion dynamics among agents, we design the decoupled interaction modules, including a Historical Interaction Encoder (HIE) and a Predictive Interaction Decoder (PID). 
After the CME, the refined trajectory tokens \( mH^T_i \) are forwarded to the subsequent HIE, then the HIE learns the historical social interaction through trajectory tokens. Following the HIE, the PID will first decode learned queries conditioned on the output of CME, then refine the queries by modeling future social interaction via ego-centric cross attention. 

Specifically, the decoder is provided with a set of learned queries that is partitioned into two groups. One group, termed the Ego Queries (denoted as \( Q_{\text{ego}}^{L} \)), is dedicated to extracting the intrinsic motion features of the primary (ego) agent, while the other group, the Contextual Queries (denoted as \( Q_{\text{ctx}}^{L} \)), is designed to capture complementary contextual information from the overall scene.
When these queries are processed by the decoder together with the unified encoder output \( H \), the decoder operation \( D(\cdot,\cdot) \) produces corresponding output tokens. Specifically, the ego queries yield the ego features
\[
Z_{\text{ego}} = D\bigl(H,\, Q_{\text{ego}}^{L}\bigr),
\]
and the contextual queries yield the contextual features
\[
Z_{\text{ctx}} = D\bigl(H,\, Q_{\text{ctx}}^{L}\bigr).
\]

The ego-centric cross-attention module is crucial for integrating the intrinsic and contextual representations. Here, the ego tokens \( Z_{\text{ego}} \) further attend to the contextual tokens \( Z_{\text{ctx}} \) using a cross-attention operation \( \text{CA}(\cdot,\cdot) \), yielding a refined ego representation:
\[
Z_{\text{ego}}^{*} = \text{CA}\bigl(Z_{\text{ego}},\, Z_{\text{ctx}}\bigr).
\]
After PID, the OmniTraj is able to not only capture the social interaction for the historical horizon, but also learn the ego-centric interaction for the future horizon, yielding more accurate trajectory prediction. Our ablation study in~\Cref{tab:ablation} validates its effectiveness.

\paragraph{Spatial-Temporal Masking and Prediction Head.}
To handle missing observations and variable sequence lengths, OmniTraj applies a spatial-temporal masking strategy that mitigates the influence of irrelevant or absent time-steps while emphasizing key fine-grained tokens essential for motion prediction. Detailed masking implementation and related ablation study can be found in our supplementary materials. 
After masking, the refined primary agent representation \(Z_{\text{prim}}^{*}\) is processed by an MLP-based prediction head to generate the future trajectory:
\[
Y_1 = \text{MLP}_{\text{pred}}\bigl(Z_{\text{prim}}^{*}\bigr).
\]

In summary, OmniTraj leverages multi-modal input embeddings to integrate trajectory, pose, and bounding box data, utilizes frame rate encoding to normalize varying temporal resolutions, and employs a robust Transformer encoder to fuse spatial and temporal features. The Transformer decoder with ego-centric cross-attention extracts both primary and contextual features through distinct learned queries, refining the primary agent’s representation via modeling the ego-centric social interaction. Finally, the spatial-temporal masking and prediction head ensure that the network is resilient to missing or inconsistent observations, resulting in accurate and generalizable trajectory prediction across diverse real-world scenarios.

\section{Experiments}
\label{sec:exp}

This section first introduces the datasets and evaluation metrics. We then evaluate our pre-training approach in zero-shot and few-shot scenarios, investigating the benefits of multi-modal learning. To further understand how different designs affect the zero-shot transfer, we do ablations of FPS-encoding and the comparison with the current state-of-the-art models in cross-domain generalization. Next, we conduct extensive performance analysis of our model on both trajectory-only and visual-cue-based datasets. Additionally, we perform an ablation study to justify our decoding designs. Finally, we analyze the model’s robustness in the two-frame prediction task. In the supplementary materials, we also show the performance on more dataset, the effect of data scaling, multi-modal inference, and implementation details.

\subsection{Datasets}
Following the approach in related work~\cite{gao2024multi}, we exclude Trajnet++~\cite{kothari2021human} and SDD~\cite{robicquet2016learning} as unseen datasets for zero-shot and few-shot evaluations. The remaining 10 datasets are used for pre-training.

We evaluate our model across several datasets to test different capabilities. For pure trajectory prediction, we use the NBA dataset~\cite{Nba2016}, notable for its intensive player interactions. Additionally, we use the pose-based JTA~\cite{fabbri2018learning} and WorldPose~\cite{jiang2024worldpose} datasets to assess whether our model can use pure trajectory to compete with previous state-of-the-art approaches that leverage 3D pose information. For robustness, we use the NuScenes~\cite{caesar2020nuscenes} to examine performance with sparse observations (two observed frames) to simulate scenarios involving suddenly appearing pedestrians. Performance on the standard ETH-UCY~\cite{pellegrini2009you,lerner2007crowds} dataset is reported in the supplementary material.

\begin{figure}[tbp]
    \centering
    \includegraphics[width=\linewidth]{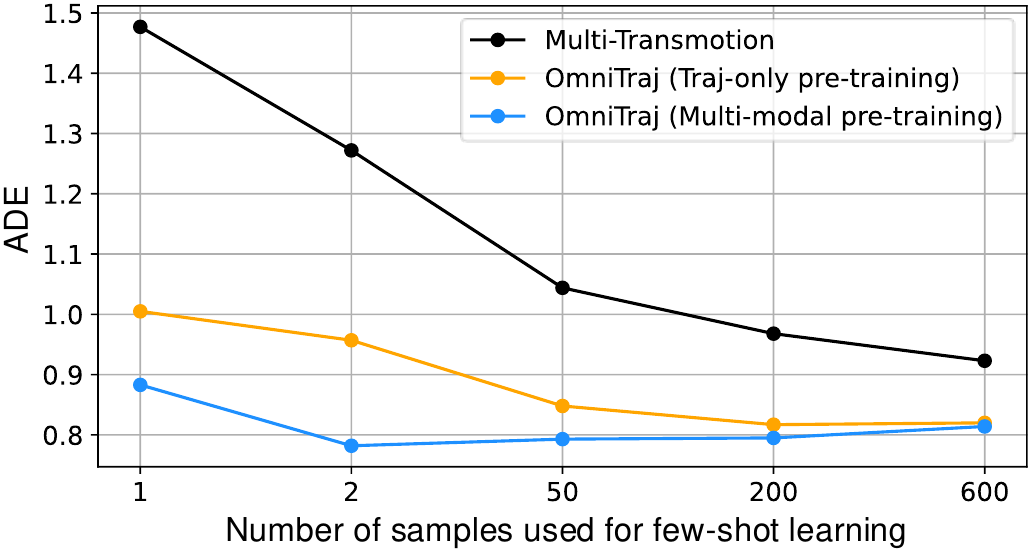}
    \caption{\textbf{Few-shot learning performance on the Trajnet++~\cite{kothari2021human} dataset.}}
    \label{fig:few-shot-trajnetpp}
\end{figure}

\begin{figure}[tbp]
    \centering
    \includegraphics[width=\linewidth]{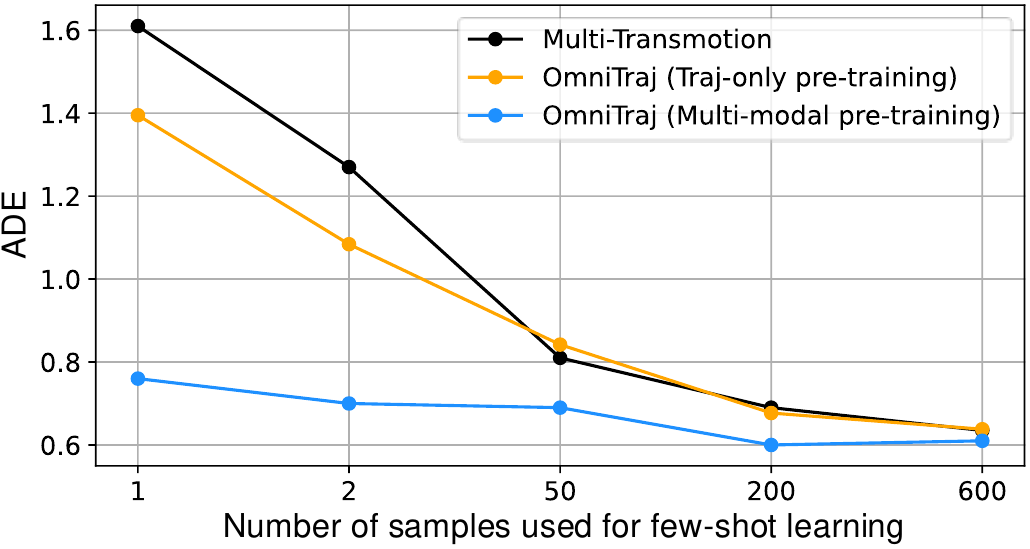}
    \caption{\textbf{Few-shot learning performance on the SDD~\cite{robicquet2016learning} dataset.}}
    \label{fig:few-shot-sdd}
\end{figure}

\subsection{Metrics}
In trajectory prediction, Average Displacement Error (ADE) and Final Displacement Error (FDE) are commonly used for deterministic prediction, while MinADE$_K$ and MinFDE$_K$ are widely adopted for selecting the best prediction from $K$ samples. Following previous works~\cite{mao2023leapfrog, gao2024multi}, we evaluate our model using MinADE$_{20}$/MinFDE$_{20}$ on the NBA dataset and ADE/FDE on the JTA and WorldPose datasets.

\subsection{Results}

\subsubsection{Zero-shot Transfer and Few-shot Learning on Unseen Data.}
In this experiment, we conduct zero-shot transfer and few-shot learning experiments on Trajnet++ and SDD, evaluating how well the learned knowledge transfers to unseen datasets. We compare our approach with Multi-Transmotion, another pre-trained model capable of being fine-tuned across different frame setups. Additionally, we investigate if a pre-trained model leveraging multi-modal data can further improve performance compared to one trained solely on pure trajectories.

To thoroughly assess generalization ability, we use different experimental setups for each dataset. Specifically, in the Trajnet++ scenario, we predict 12 future frames given 9 historical frames, whereas in SDD, we predict 12 future frames given only 6 historical frames.

\Cref{fig:few-shot-trajnetpp} and~\Cref{fig:few-shot-sdd} illustrate the learning curves under extremely limited data conditions. Our multi-modal pre-training consistently generalizes better than trajectory-only pre-training. This suggests that it has learned more generalizable knowledge, enabling it to predict trajectories in unseen datasets regardless of frame setups.

\Cref{tab:zero-shot} quantitatively presents the zero-shot performance of all the current available pre-trained models capable of handling different frame setups. Notably, our method surpasses Multi-Transmotion by over 70\%, highlighting its practicality for unseen scenarios. In few-shot learning, our model outperforms Multi-Transmotion fine-tuned with 200 samples using only 2 training samples, demonstrating its strong generalization capability.

\begin{table}[tbp]
    \centering
    \resizebox{0.5\textwidth}{!}{

    \begin{tabular}{lccc}
    \toprule
        \textbf{Models}  &\textbf{Trajnet++} & \textbf{SDD} \\ 
        {} & ADE (gain) & ADE (gain) \\
        \midrule
        Multi-Transmotion~\cite{gao2024multi} & 3.40  & 3.58 \\ 
        OmniTraj (Traj-only pre-training) & 1.57 (53.8\%) & 1.91 (46.6\%)\\
        OmniTraj (Multi-modal pre-training)& \textbf{1.01, (70.2\%)} & \textbf{0.93, (74.0\%)} \\ 
        \bottomrule
    \end{tabular}
    }
    \caption{\textbf{Zero-shot transfer on two datasets with different frame setups.}}

    \label{tab:zero-shot}

\end{table}

\subsubsection{Comparisons and Ablations on zero-shot transfer.}

To rigorously evaluate a model's ability to generalize to unseen temporal dynamics, we design a challenging zero-shot cross-setup study using the NBA dataset. We partition the training data into two subsets with distinct temporal properties:

\begin{itemize} 
    \item Setup 1: Predict 20 future frames given 10 observed frames at 5 FPS.
    \item Setup 2: Predict 8 future frames given 4 observed frames at 2.5 FPS.
\end{itemize}

We then evaluate all models on a completely unseen test set, Setup 3, which requires predicting 3 future frames from 3 observed frames at 1 FPS.

For this study, we compare OmniTraj against two state-of-the-art baselines that represent distinct strategies for zero-shot transfer in trajectory prediction: a data-unaware discrete model (Multi-Transmotion) and a continuous-time model (TrajSDE). All models are trained on a mixture of Setup 1 and 2 and evaluated directly on Setup 3, providing a study of zero-shot transfer.

\begin{table}[tbp]
\centering
\resizebox{0.5\textwidth}{!}{
\begin{tabular}{lc}
\toprule
\textbf{Methods} & \textbf{MinADE$_{20}$/MinFDE$_{20}$} \\

\midrule

Multi-Transmotion ~\cite{gao2024multi} & 1.68/2.15 \\
TrajSDE ~\cite{park2024improving} & 1.71/2.05 \\
\midrule
OmniTraj (Ours) and Ablations:	& {} \\
w/o FPS-encoder & 1.87/2.49 \\
w/ Film-based~\cite{perez2018film} FPS-encoder & 1.62/2.26 \\
w/ Codebook FPS-encoder & 1.46/1.66 \\
w/ MLP FPS-encoder, Concat. as extra token & 1.74/2.36 \\
w/ MLP FPS-encoder, Latent-space summation & \textbf{1.18/1.22} \\

\bottomrule

\end{tabular}
}
\caption{\textbf{Comparison and ablations on zero-shot transfer.}}

\label{tab:fps_encoder_design}

\end{table}

The results in Table~\ref{tab:fps_encoder_design} provide several key insights. First, they confirm that zero-shot transfer is a significant challenge; the Multi-Transmotion and our model without an FPS-encoder both fail to adapt to the unseen 1 FPS setup. Second, while introducing any form of FPS-awareness improves performance, the design is critical. Our proposed MLP-encoder with latent-space summation is the most effective design, reducing MinADE$_{20}$ by 36.9\% and MinFDE$_{20}$ by 51.0\% compared to the version without any FPS-encoder. Most importantly, our simple and efficient approach not only outperforms its own more complex variants (e.g., FiLM-based) but also the continuous-time model (TrajSDE). This strongly supports that a direct and explicit signal of data context is a more robust and effective solution than adopting complex internal dynamic models.

\subsubsection{Performance on Trajectory-only Dataset.}

The NBA dataset is a highly interactive dataset, as basketball is a team-based activity that requires the model to effectively understand human motion dynamics and social interactions between agents. Following prior work~\cite{mao2023leapfrog, gao2024multi}, we generate multiple predictions using 10 historical frames to predict 20 future frames at 5 FPS. As shown in Table~\ref{tab:result_nba}, our method consistently outperforms all previous baselines, achieving a new state-of-the-art on this dataset. This performance can be attributed to both our architectural design and pre-trained knowledge, as validated in the later ablation study.

\begin{figure*}
    \centering
    \subfloat[Prediction from Multi-Transmotion~\cite{gao2024multi}]{%
        \includegraphics[width=\textwidth]{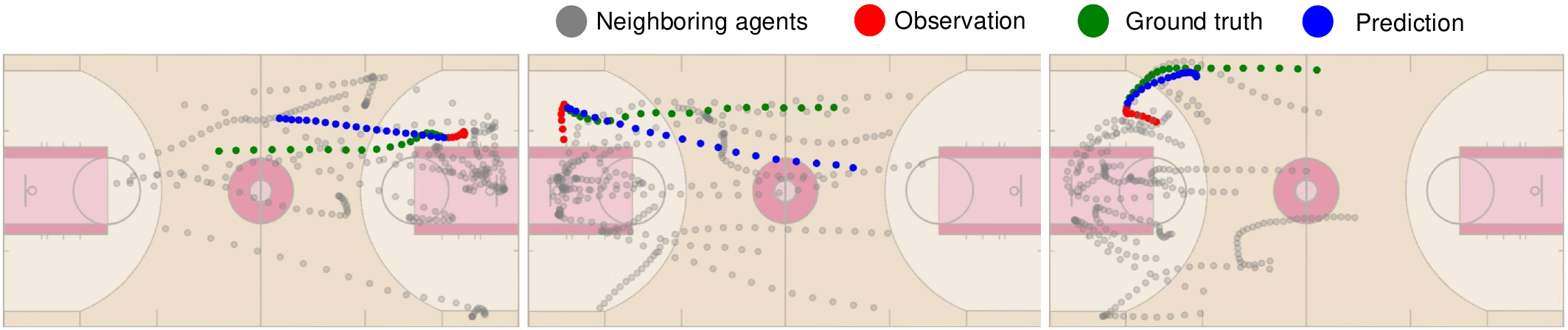}
    }
    \vfill
    \subfloat[Prediction from our OmniTraj]{%
        \includegraphics[width=\textwidth]{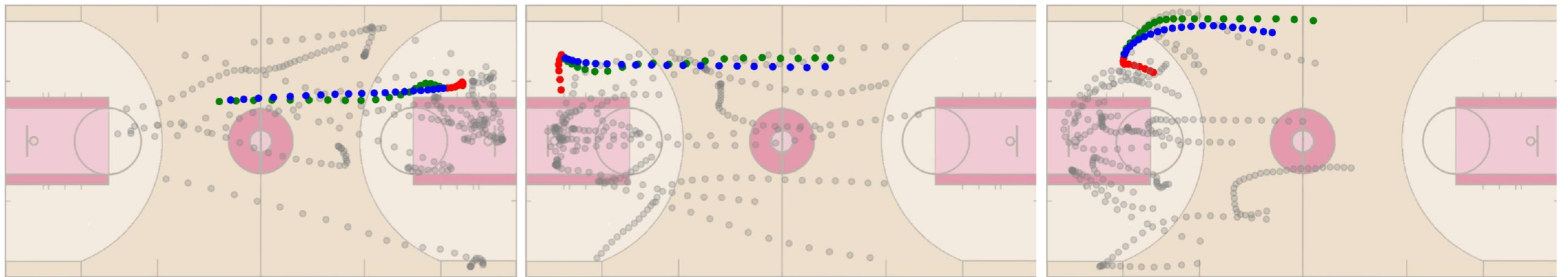}
    }
    \caption{\textbf{Qualitative results on the NBA~\cite{Nba2016} dataset.}}
    \label{fig:vis_nba}
\end{figure*}

Figure~\ref{fig:vis_nba} presents a qualitative comparison between our model and the previous state-of-the-art, Multi-Transmotion, demonstrating our model's superior accuracy.
For example, in the middle case, our model accurately predicts the trajectory despite the ego-agent making a sharp turn in the last observed frames. In contrast, Multi-Transmotion exhibits greater deviation from the ground truth, accumulating more prediction error over time.

\begin{table}[tbp]
    \centering
    \resizebox{0.5\textwidth}{!}{

    \begin{tabular}{lcc}
    \toprule
        \textbf{Models}  & \textbf{Venue} & \textbf{MinADE$_{20}$/MinFDE$_{20}$} \\ 
        \midrule
        Social-LSTM~\cite{alahi2016social} & CVPR 16 & 1.65/2.98 \\ 
        Social-GAN~\cite{gupta2018social} & CVPR 18  & 1.59/2.41 \\ 
        STGAT~\cite{huang2019stgat} & ICCV 19  & 1.40/2.18 \\
        Social-STGCNN~\cite{mohamed2020social} & ICCV 20  & 1.53/2.26 \\ 
        STAR~\cite{yu2020spatio} & ECCV 20 & 1.13/2.01 \\ 
        Trajectron++~\cite{salzmann2020trajectron++} & ECCV 20  & 1.15/1.57 \\ 
        MemoNet~\cite{xu2022remember} & CVPR 22 & 1.25/1.47 \\ 
        GroupNet~\cite{xu2022groupnet} & CVPR 22  & 0.96/1.30 \\
        MID~\cite{gu2022stochastic} & CVPR 22  & 0.96/1.27 \\
        LED~\cite{mao2023leapfrog} & CVPR 23  & 0.81/1.10 \\
        Social-Transmotion~\cite{saadatnejad2024socialtransmotion} & ICLR 24  & 0.78/1.01 \\ 
        Multi-Transmotion~\cite{gao2024multi} & CoRL 24  & 0.75/0.97 \\ 
        NMRF~\cite{fang2025neuralized} & ICLR 25 & 0.75/0.97 \\
        OmniTraj & -- & \textbf{0.73/0.91} \\
        \bottomrule
    \end{tabular}
    }
        \caption{\textbf{Quantitative results on the NBA~\cite{Nba2016} dataset.}}

    \label{tab:result_nba}

\end{table}

\subsubsection{Challenge Pose-based Methods by Using Trajectory-only.}
\begin{table}[tbp]
    \centering
    \vspace{-5pt}
     
    \resizebox{.5\textwidth}{!}{
    \begin{tabular}{lcc}
    \toprule
        \textbf{Models}  & \textbf{Inference modality} &\textbf{ADE/FDE} \\
        \midrule
        Vanilla-LSTM~\cite{alahi2016social}& T & 1.44/3.25\\
        Social-LSTM~\cite{alahi2016social} & T & 1.21/2.54  \\
        Trajectron++~\cite{salzmann2020trajectron++}  & T & 1.18/2.53 \\

        Autobots~\cite{girgis2022latent}  & T &  1.20/2.70 \\
        EqMotion~\cite{xu2023eqmotion}  & T & 1.13/2.39 \\
        Social-Transmotion~\cite{saadatnejad2024socialtransmotion}   & T + 3D P& 0.94/1.94\\
        Multi-Transmotion~\cite{gao2024multi}   & T + 3D P& 0.91/1.89\\
        EmLoco~\cite{taketsugu2025physical} & T + 3D P & 0.97/1.91 \\
        OmniTraj & T & \textbf{0.90/1.81}\\

        \bottomrule
    \end{tabular}
    }

         \caption{\textbf{Quantitative results on the JTA~\cite{fabbri2018learning} dataset.} Social-Transmotion~\cite{saadatnejad2024socialtransmotion}, Multi-Transmotion~\cite{gao2024multi} and EmLoco~\cite{taketsugu2025physical} are trained on Trajectory and 3D Pose modalities since they can leverage pose. (`T' and `P' abbreviate Trajectory and Pose keypoints).}

\label{tab:jta}

\end{table}

Previous studies~\cite{saadatnejad2024socialtransmotion} have demonstrated that 3D pose keypoints can enhance trajectory prediction performance. However, accurately capturing visual cues such as 3D pose is challenging in real-world scenarios. Given this disparity, we want to see if our pre-trained model can compete with the pose-based models when we only feed trajectories to our model.

\Cref{tab:jta} presents the quantitative results on JTA. We compare OmniTraj against both pose-based and trajectory-only models. Leveraging large-scale pre-training, our method achieves state-of-the-art performance using only trajectory data, surpassing all previous baselines. Additionally, we evaluate our model and some selected baselines on the WorldPose dataset, specifically in a football scenario. As shown in \Cref{tab:result_worldpose}, our method consistently outperforms both pose-based and trajectory-only models.

\begin{table}[tbp]
    \centering
    \resizebox{0.5\textwidth}{!}{

    \begin{tabular}{lccc}
    \toprule
        \textbf{Models}  & Input modality &\textbf{ADE/FDE} \\ 
        \midrule
        Social-LSTM~\cite{alahi2016social} &T&  2.66/6.30 \\ 
        Directional-LSTM~\cite{kothari2021human}  &T&  2.62/6.25 \\ 
        Autobots~\cite{girgis2022latent}  & T& 2.84/5.80 \\
        Social-Transmotion~\cite{saadatnejad2024socialtransmotion}  &T& 2.68/6.11 \\ 
        Social-Transmotion~\cite{saadatnejad2024socialtransmotion} &T+3d P& 2.43/5.73 \\ 
        OmniTraj & T & \textbf{2.38/5.46} \\
        \bottomrule
    \end{tabular}
    }

        \caption{\textbf{Quantitative results on the WorldPose~\cite{jiang2024worldpose} dataset.} Similarly, `T' and `P' abbreviate Trajectory and Pose keypoints.}

    \label{tab:result_worldpose}

\end{table}

\subsubsection{Ablations on the Decoupled interaction modules.}
We conduct a detailed ablation study to analyze the contribution of each component within our decoupled interaction modules. As detailed in Table~\ref{tab:ablation}, our analysis begins with a baseline using only the backbone network, without any interaction modules. The results show that incorporating the Historical Interaction Encoder (HIE) alone provides a substantial performance boost. Similarly, adding the Predictive Interactive Decoder (PID), which consists of the decoder and cross-attention, also markedly enhances performance. The best results are achieved when all modules are enabled, allowing the model to process historical and future predictive interactions separately. This outcome validates the effectiveness of our decoupled design.

\begin{table}[!tbp]
    \centering
    \resizebox{0.4\textwidth}{!}{

    \begin{tabular}{lccc}
    \toprule
        \textbf{HIE}  &\textbf{Decoder} & \textbf{Cross-attention }&\textbf{MinADE$_{20}$/MinFDE$_{20}$} \\ 
        \midrule
         -- & --  & -- & 0.91/1.33\\
        
        \CheckmarkBold &  -- & -- & 0.77/0.98 \\
        \CheckmarkBold & \CheckmarkBold & -- & 0.75/0.97\\
          -- & \CheckmarkBold & \CheckmarkBold & 0.80/1.04\\
        \CheckmarkBold & \CheckmarkBold & \CheckmarkBold & \textbf{0.74/0.94}\\ 
        \bottomrule
    \end{tabular}
    }

        \caption{\textbf{Ablation study on the Decoupled interaction Modules.} Models are trained from scratch and evaluated on the entire test set of the NBA dataset.}

    \label{tab:ablation}

\end{table}

\subsubsection{Robustness on Two-frame Prediction.}
In road traffic scenarios, pedestrians may suddenly appear, limiting the amount of historical information available for prediction. Thus, it is important to examine how performance degrades when the model has access to only two historical frames.

As shown in~\Cref{tab:two-frame}, we compare OmniTraj with a Social-Transmotion~\cite{saadatnejad2024socialtransmotion} model specifically trained on the NuScenes setup, which serves as our baseline. When only two frames are available, the specific model experiences a 57\% drop in ADE, while our pre-trained model remains highly robust, with only an 11\% drop in ADE and an 8\% drop in FDE. These results highlight the practical applicability of our approach in real-world scenarios with limited historical observations.

\begin{table}[!tbp]
    \centering
    \resizebox{0.4\textwidth}{!}{

    \begin{tabular}{lcc}
    \toprule
        \textbf{Models}  &\textbf{All frames used} & \textbf{Two-frame only} \\ 
        {} & ADE/FDE & ADE/FDE\\
        \midrule
        Social-Transmotion & 0.69/1.45  & 1.08/2.03\\ 
        OmniTraj & \textbf{0.66/1.37} & \textbf{0.73/1.48} \\ 
        \bottomrule
    \end{tabular}
    }

        \caption{\textbf{Two-frame prediction on the Nuscenes~\cite{caesar2020nuscenes} dataset.}}

    \label{tab:two-frame}

\end{table}

\section{Conclusion}
\label{sec:conclusion}

We tackled the critical challenge of zero-shot transfer in human trajectory prediction. Our systematic investigation revealed that a simple, explicit frame-rate conditioning mechanism is a more effective solution than current data-unaware or continuous-time models. Our model, OmniTraj, leverages this insight to achieve state-of-the-art zero-shot performance, reducing prediction error by over 70\% on unseen datasets. This work validates a simple yet powerful approach for creating more robust and practical forecasting models.

\newpage
\section{Acknowledgement}

The authors would like to thank Zimin Xia, Mohamed Abdelfattah and Bastien Van Delft for their valuable feedback. This work was supported by Sportradar\footnote{\href{https://sportradar.com/}{https://sportradar.com/}} (Yang's Ph.D.), Honda R\&D Co., Ltd, and Valeo Paris.

\bibliography{main}

\newpage
\twocolumn[
\begin{center}
    {\huge \bfseries Supplementary Materials}
\end{center}
\vspace{1.5cm}
]

In supplementary materials, we begin by detailing the proposed UniHuMotion++ data framework, a large-scale collection that unifies 12 diverse human motion datasets totaling 859 hours of data. We follow this with a description of the data splitting methodology used across all datasets for pre-training.

Subsequently, we present additional experimental results on the ETH-UCY benchmark to demonstrate our model's competitive performance. After that, we analyze the effect of data scaling. The document also includes ablation studies that validate our design choices by the temporal masking ratio and multi-modal fusion at inference. Finally, we provide comprehensive implementation details covering model architecture and training parameters.

\section{Our Proposed  Unified Human Motion Data Framework (UniHuMotion++)}
The detailed information about our UniHuMotion++ framework is listed in \Cref{tab:UniHuMotion}. This framework contains 859 hours of human motion data from 12 diverse datasets, which makes it a large-scale, multi-modal dataset invariant to frame rate and horizon length. With the raw data, the data framework is flexible to accommodate different temporal setups with specified modalities.

\begin{table*}[htbp]

    \centering
     \begin{adjustbox}{max width=\textwidth}
    \begin{tabular}{lcccccccccc}
    \toprule
        \textbf{Dataset} & \textbf{R(eal)/S(ynt.)} & \textbf{Applied FPS} & \textbf{Obs. frames} & \textbf{Pred. frames} & \textbf{Data size} & \textbf{Traj.} & \textbf{3D BB} & \textbf{2D BB} & \textbf{3D Pose} & \textbf{2D Pose}\\
        \midrule
        NBA SportVU~\cite{Nba2016} & R & 5 & 10 & 20 & 150h & \CheckmarkBold & \XSolidBrush & \XSolidBrush & \XSolidBrush & \XSolidBrush \\
        JRDB-Pose~\cite{vendrow2023jrdbpose} & R & 2.5 & 9 & 12 & 1.1h & \CheckmarkBold & \CheckmarkBold & \CheckmarkBold & \XSolidBrush & \CheckmarkBold \\
        JTA~\cite{fabbri2018learning} & S & 2.5 & 9 & 12 & 4.3h & \CheckmarkBold & \CheckmarkBold & \CheckmarkBold & \CheckmarkBold & \CheckmarkBold \\
        Human3.6M~\cite{ionescu2013human3} & R & 5 & 10 & 20 & 20h & \CheckmarkBold & \CheckmarkBold & \XSolidBrush & \CheckmarkBold & \XSolidBrush \\
        AMASS~\cite{mahmood2019amass} & R & 25 & 50 & 25 & 42h & \CheckmarkBold & \CheckmarkBold & \XSolidBrush & \CheckmarkBold & \XSolidBrush \\
        3DPW~\cite{von2018recovering} & R & 25 & 50 & 25 & 0.5h & \CheckmarkBold & \CheckmarkBold & \XSolidBrush & \CheckmarkBold & \XSolidBrush \\
        Nuscenes~\cite{caesar2020nuscenes} & R & 2 & 4 & 12 & 5.5h & \CheckmarkBold & \XSolidBrush & \XSolidBrush & \XSolidBrush & \XSolidBrush \\
        WOMD~\cite{ettinger2021large} & R & 10 & 20 & 60 & 570h & \CheckmarkBold & \CheckmarkBold & \XSolidBrush & \XSolidBrush & \XSolidBrush \\
        Argoverse2~\cite{wilson2023argoverse} & R & 10 & 20 & 60 & 4.2h & \CheckmarkBold & \CheckmarkBold & \XSolidBrush & \XSolidBrush & \XSolidBrush \\
        WorldPose~\cite{jiang2024worldpose} & R & 5 & 10 & 20 & 0.8h & \CheckmarkBold & \CheckmarkBold & \XSolidBrush & \CheckmarkBold & \XSolidBrush \\
        SDD~\cite{robicquet2016learning} & R & 2.5 & 6 & 12 & 8.6h & \CheckmarkBold & \XSolidBrush & \XSolidBrush & \XSolidBrush & \XSolidBrush \\
        Trajnet++~\cite{kothari2021human} & R+S & 2.5 & 9 & 12 & 52h & \CheckmarkBold & \XSolidBrush & \XSolidBrush & \XSolidBrush & \XSolidBrush \\

        \bottomrule
    \end{tabular}
    \end{adjustbox}

         \caption{\textbf{UniHuMotion++: A Unified Human Motion Data Framework.}}

\label{tab:UniHuMotion}
\vspace{-5pt}
\end{table*}

\section{Data Splits of All Datasets in UniHuMotion++ Framework}
\label{sec:data_splits}

For pre-training, we followed Social-Transmotion~\cite{saadatnejad2024socialtransmotion} to prepare data splits for JRDB-Pose~\cite{vendrow2023jrdbpose} and JTA~\cite{fabbri2018learning} and adopted the Multi-Transmotion~\cite{gao2024multi} methodology to split NBA SportVU~\cite{Nba2016}, Human3.6M~\cite{ionescu2013human3}, AMASS~\cite{mahmood2019amass}, and 3DPW~\cite{von2018recovering}. Similarly, we followed ScenarioNet~\cite{li2023scenarionet} for NuScenes~\cite{caesar2020nuscenes}, WOMD~\cite{sun2020scalability}, and Argoverse2~\cite{wilson2023argoverse}. The WorldPose~\cite{jiang2024worldpose}, trajnet++~\cite{kothari2021human}, and SDD~\cite{robicquet2016learning} dataset were prepared using its official splits.

\section{Results on ETH-UCY}
\label{sec:eth_ucy}

We also fine-tune our model on another trajectory-only dataset, ETH-UCY~\cite{pellegrini2009you,lerner2007crowds}, following prior works\cite{xu2023eqmotion,salzmann2020trajectron++} to predict 12 future frames based on 8 past frames at 2.5 FPS. \Cref{tab:eth_ucy} presents the quantitative results, showing that our model achieves competitive performance compared to previous state-of-the-art methods, particularly on the Zara1 and Zara2 subsets, further demonstrating its strong generalization ability.

\begin{table*}[tbp]
    \centering
    \resizebox{0.8\textwidth}{!}{

    \begin{tabular}{lcccccc}
    \toprule
        \textbf{Models}  & \textbf{ETH} &\textbf{Hotel} & \textbf{Univ} & \textbf{Zara1} & \textbf{Zara2} & \textbf{Average} \\ 
        \midrule
        Social-GAN~\cite{gupta2018social} & 0.87/1.62&  0.67/1.37 & 0.76/1.52 & 0.35/0.68 & 0.42/0.84 & 0.61/1/21 \\ 
        STGCNN~\cite{mohamed2020social} & 0.64/1.11 & 0.49/0.85 & 0.44/0.79 & 0.34/0.53 & 0.30/0.48 & 0.44/0.75\\
        Trajectron++~\cite{salzmann2020trajectron++} & 0.61/1.03 & 0.20/0.28 & 0.30/0.55 & 0.24/0.41 & 0.18/0.32 & 0.31/0.52 \\
        GroupNet~\cite{xu2022groupnet} & 0.46/0.73 & 0.15/0.25 & 0.26/0.49 & 0.21/0.39 & 0.17/0.33 & 0.25/0.44 \\
        AgentFormer~\cite{yuan2021agentformer} & 0.46/0.80 & 0.14/0.22 & 0.25/0.45 & \underline{0.18}/\textbf{0.30} & \underline{0.14}/\underline{0.24} & \underline{0.23}/0.40\\
        GP-Graph~\cite{bae2022learning} & 0.43/0.63 & 0.18/0.30 & \underline{0.24}/\underline{0.42} & \textbf{0.17}/\underline{0.31} & 0.15/0.29 & \underline{0.23}/0.39 \\
        NSPN~\cite{bae2022non} & \underline{0.36}/\underline{0.59} & 0.16/0.25 & \textbf{0.23}/\textbf{0.39} & \underline{0.18}/0.32 & \underline{0.14}/0.25 & \textbf{0.21}/0.36\\
        EqMotion~\cite{xu2023eqmotion} & 0.40/0.61 & \textbf{0.12}/\textbf{0.18} & \textbf{0.23}/0.43 & \underline{0.18}/0.32 & \textbf{0.13}/\textbf{0.23} & \textbf{0.21}/\underline{0.35} \\
        SingularTrajectory~\cite{bae2024singulartrajectory} & \textbf{0.35}/\textbf{0.42} & \underline{0.13}/\underline{0.19} & 0.25/0.44 & 0.19/0.32 & 0.15/0.25 & \textbf{0.21}/\textbf{0.32} \\
        OmniTraj & 0.38/\underline{0.59} & 0.14/0.20 & \textbf{0.23}/\underline{0.42} & \textbf{0.17}/\textbf{0.30} & \textbf{0.13}/\textbf{0.23} & \textbf{0.21}/\underline{0.35}\\
        \bottomrule
    \end{tabular}
    }

        \caption{\textbf{Quantitative results on the ETH-UCY~\cite{pellegrini2009you,lerner2007crowds} dataset.} MinADE$_{20}$/MinFDE$_{20}$ (meters) are reported, with the \textbf{Best} numbers bolded and \underline{second-best} numbers underlined.}

    \label{tab:eth_ucy}

\end{table*}

\section{Data Scaling Effect}
\begin{figure}[tbp]
    \centering
    \includegraphics[width=\linewidth]{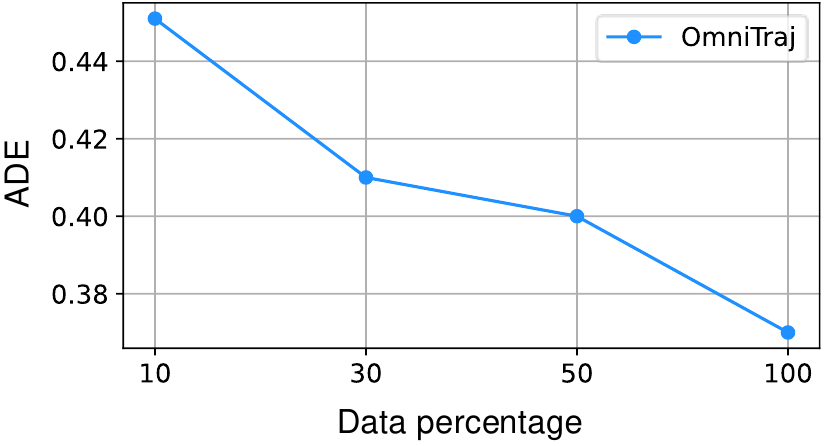}
    \caption{\textbf{Relationship between performance and data percentage used in pre-training.}}
    \label{fig:data-scaling}
\end{figure}

To examine the data scalability, we pre-train our model on different data proportions and evaluate its validation performance. This analysis examines how the model benefits from additional training data and whether performance gains persist. As shown in \Cref{fig:data-scaling}, the model's performance consistently improves with larger datasets, demonstrating enhanced generalization across datasets with different frame setups and motion dynamics.

\section{Implementation Details}
\label{sec:implementation}

The cross-modality encoder consists of six layers with four attention heads, while the historical interaction encoder has four layers with four heads. Regarding the predictive interaction decoder, the Transformer decoder contains two layers with four heads, followed by a cross-attention layer. For optimization, we use the Adam optimizer~\citep{kingma2014adam} with an initial learning rate of $1 \times 10^{-4}$, which decays by a factor of 0.1 after 80\% of the total 30 epochs. Pre-training is conducted on six NVIDIA H100 GPUs, each with 80GB of memory. For supervision, we use L2 loss to guide the predicted trajectory outputs.

Regarding masking strategies, we adopt a 30\% modality masking~\cite{saadatnejad2024socialtransmotion} to learn cross-modality representation and a 50\% spatial masking on human body pose~\cite{yuan2023hap}. Temporally, we retain the last two observed frames and apply a 75\% masking ratio on the other observed frames, the justification is given in ~\Cref{tab:abl_masking}. 

\section{Model size and inference speed}
OmniTraj has approximately 7.5 million parameters. We evaluate the inference speed on the JTA~\cite{fabbri2018learning} dataset. The inference speed is 595$\pm$19 predictions/sec on a single NVIDIA H100 GPU, measured over five runs with around 5k samples each.

\section{Ablation on Temporal Masking Ratio.}
To empirically validate how different temporal masking ratios affect the performance in our task, we conducted an ablation study on a large subset of the NBA dataset ($>$130k samples), varying the masking ratio as shown in \Cref{tab:abl_masking}. 
The results confirm that a 75\% temporal masking ratio yields slightly better performance compared to other tested ratios in our task.

\begin{table}[!tbp]
    \centering
    \resizebox{0.45\textwidth}{!}{

    \begin{tabular}{lccccc}
    \toprule
    {} & 10\% & 25\% & 50\% & 75\% & 90\%\\
    \midrule
    MinADE$_{20}$ & 0.9253 & 0.9247 & 0.9222 & 0.9171 & 0.9427 \\
        
    \bottomrule
    \end{tabular}
    }
    \caption{\textbf{How different temporal masking ratios affect the performance.}}
    \label{tab:abl_masking}

\end{table}

\section{Effectiveness of multi-modal fusion and inference-time adaptation.}
OmniTraj utilizes its Cross-Modality Encoder (CME), where attention mechanisms learn to fuse and weigh features contextually. The multi-modal pre-training includes modality masking to prevent over-reliance on any single modality. We conduct both multi-modal training and multi-modal inference on JTA dataset, validating our approach in \Cref{tab:multi_modal_inference}: adding the 3D BBox (B), or 3D Pose (P), or both of them consistently improves performance, demonstrating the CME's effective multi-modal fusion.

\begin{table}[htbp]
    \centering

    \begin{tabular}{llc}
    \toprule
     & Inference Modality& ADE/FDE \\
    \midrule
    OmniTraj & T & 0.94/1.88 \\
    OmniTraj & T + 3d B & 0.91/1.81\\
    OmniTraj & T + 3d P & 0.87/1.74\\
    OmniTraj & T + 3d B + 3d P & 0.86/1.73\\
        
    \bottomrule
    \end{tabular}

    \caption{\textbf{Multi-modal inference on JTA dataset.} `T', `B', and `P' abbreviate Trajectory Bounding boxes, and Pose keypoints)}
    \label{tab:multi_modal_inference}

\end{table}
\end{document}